\documentclass[10pt,twocolumn,letterpaper]{article}

\usepackage{cvpr}              % To produce the CAMERA-READY version
\definecolor{cvprblue}{rgb}{0.21,0.49,0.74}
\usepackage[pagebackref,breaklinks,colorlinks,allcolors=cvprblue]{hyperref}

\usepackage{graphicx}
\usepackage{amsmath}
\usepackage{amssymb}
\usepackage{booktabs}
\usepackage{multirow} 
\usepackage{adjustbox} 
\usepackage{tabularx}
\usepackage[accsupp]{axessibility}

\def\paperID{} % *** Enter the Paper ID here
\def\confName{CVPR}
\def\confYear{2026}

\title{MaskDiME: Adaptive Masked Diffusion for Precise and Efficient Visual Counterfactual Explanations}

\author{
Changlu Guo \quad
Anders Nymark Christensen \quad
Anders Bjorholm Dahl \quad
Morten Rieger Hannemose\\
Technical University of Denmark, Denmark\\
{\tt\small \{chagu,anym,abda,mohan\}@dtu.dk}
}

\begin{document}
\maketitle
\begin{abstract}
Visual counterfactual explanations aim to reveal the minimal semantic modifications that can alter a model’s prediction, providing causal and interpretable insights into deep neural networks. However, existing diffusion-based counterfactual generation methods are often computationally expensive, slow to sample, and imprecise in localizing the modified regions. To address these limitations, we propose MaskDiME, a simple, fast, yet effective diffusion framework that unifies semantic consistency and spatial precision through localized sampling. Our approach adaptively focuses on decision-relevant regions to achieve localized and semantically consistent counterfactual generation while preserving high image fidelity. Our training-free framework, MaskDiME, performs inference over \textbf{30$\times$} faster than the baseline and achieves comparable or state-of-the-art performance across five benchmark datasets spanning diverse visual domains, establishing a practical and generalizable solution for efficient counterfactual explanation. The project page is available at \href{https://clguo.github.io/MaskDiME/}{https://clguo.github.io/MaskDiME/}.
\end{abstract}    
\section{Introduction}

\begin{figure}[t]
  \centering
   \includegraphics[width=1\linewidth]{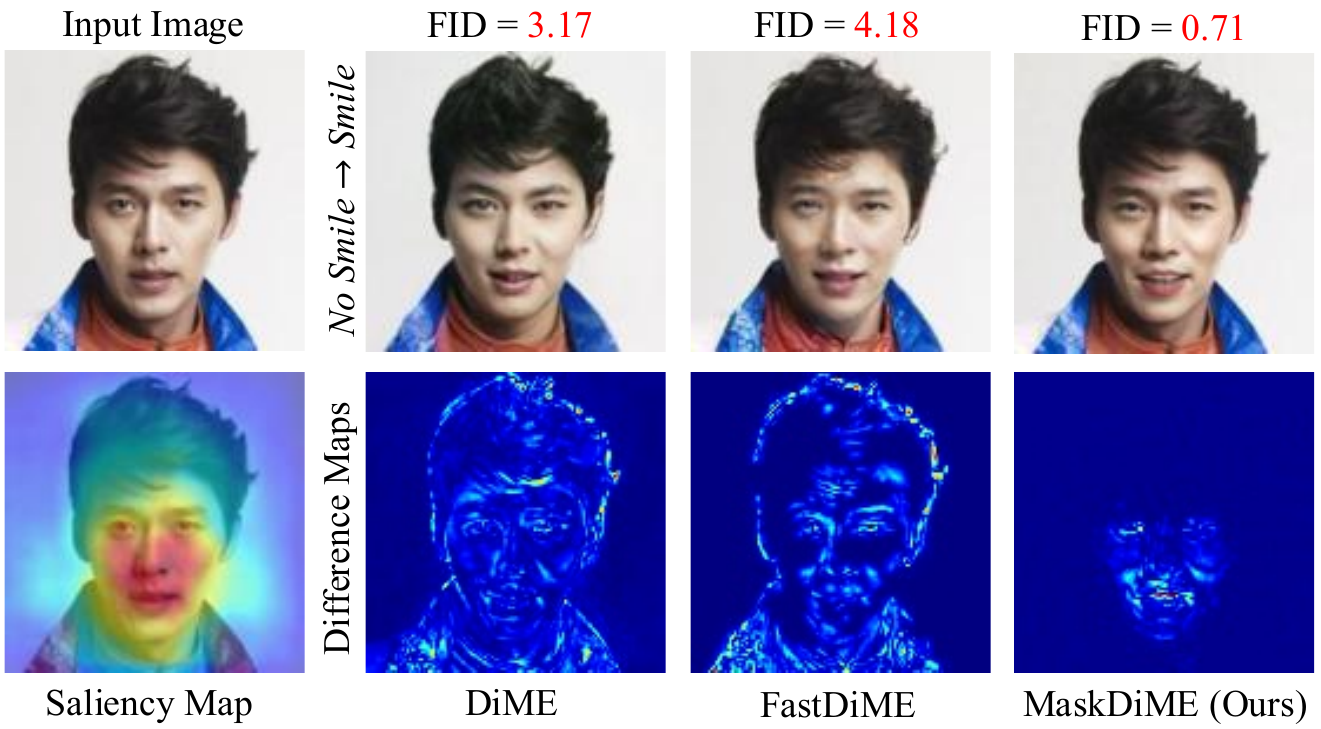}

\caption{Previous methods such as DiME~\cite{jeanneret2022diffusion} and FastDiME~\cite{weng2024fast} 
produce global or scattered edits, whereas our \textbf{MaskDiME} achieves localized, 
decision-relevant modifications consistent with the classifier's saliency map, computed via SmoothGrad~\cite{smilkov2017smoothgrad}.}

   \label{fig:teaser}
\end{figure}

For Explainable Artificial Intelligence (XAI), it is important to understand \emph{why} a model makes a particular decision. Using counterfactuals for XAI of images aims to alter an image while retaining its overall content and changing the appearance of specific attributes, such as turning a person from not smiling to smiling, as illustrated in \cref{fig:teaser}. 
This idea forms the basis for Visual Counterfactual Explanations (VCEs), which aim to generate visually realistic counterfactuals through minimal and spatially precise modifications that change the model prediction. Unlike saliency maps that only highlight correlated regions, VCEs align with human reasoning by answering the question:\emph{ “What must change in this image for the model to decide differently?”}

Although the concept of VCEs is intuitive, generating high-quality counterfactuals remains a challenging task. 
% An ideal VCE should satisfy key criteria: \textit{validity} (flipping the model’s prediction), \textit{realism} (staying on the natural image manifold), \textit{sparsity and proximity} (making minimal, localized changes), and \textit{diversity} (offering multiple plausible alternatives to reveal distinct causal factors)~\cite{jeanneret2022diffusion,jeanneret2023adversarial}.
Recently, diffusion models~\cite{ho2020denoising} have demonstrated remarkable ability in producing visually realistic and semantically coherent images, inspiring a series of diffusion-based counterfactual approaches~\cite{jeanneret2022diffusion,jeanneret2023adversarial,weng2024fast,sobieski2024rethinking}. Despite these advances, existing approaches still face two major challenges. 
First, the computational cost remains high.
Most diffusion-based counterfactual methods rely on nested denoising, multi-stage processing, or large latent diffusion architectures, resulting in slow inference and large memory usage~\cite{jeanneret2022diffusion,jeanneret2023adversarial,farid2023latent,yeganeh2025latent}. 
Although recent work~\cite{weng2024fast} improves efficiency, the gains in speed come with limited generative quality, revealing a second major challenge—the lack of spatial and semantic precision. Most existing methods rely on global classifier guidance or implicit conditioning~\cite{jeanneret2022diffusion,weng2024fast,farid2023latent,jeanneret2024text,yeganeh2025latent}, leading to indiscriminate signal propagation and spatially imprecise, semantically entangled edits that may change the overall appearance of the image and make it difficult to tell which parts of the image explain the decision (see \cref{fig:teaser}).

While spatial-constraint methods improve localization, their static masks remain fixed during reverse diffusion and cannot effectively adapt to evolving semantic regions, such as changes in facial expressions, resulting in imprecise edits and limited generalization.

To address these limitations, we revisit the interaction between classifier guidance and the diffusion trajectory. We argue that enabling the model to adaptively focus on decision-relevant regions during reverse diffusion is key to achieving precise and semantically coherent counterfactual explanations. Building upon this insight, we propose MaskDiME—a simple yet effective, training-free diffusion framework that unifies semantic consistency and spatial precision in visual counterfactual generation. MaskDiME introduces an adaptive dual-mask mechanism that utilizes two masks derived from classifier gradients, dynamically constraining updates to decision-critical regions throughout the reverse diffusion process. This spatially adaptive guidance enables minimal yet semantically consistent image edits, and even with one-step clean image estimation, maintains efficiency while preserving high image fidelity. Our main contributions are as follows: (1) We present a new perspective that highlights the role of the interaction between classifier guidance and diffusion dynamics in shaping spatial precision and semantic consistency; (2) We propose a spatially adaptive diffusion framework, MaskDiME, which transforms traditional global updates into decision-driven local edits, achieving significant improvements in both efficiency and precision; (3) We design an adaptive dual-mask mechanism that dynamically identifies and constrains decision-relevant regions at each diffusion step, ensuring semantically consistent local edits;      (4) We validate the efficiency and generalization of MaskDiME, achieving much faster inference without extra training or fine-tuning, while maintaining superior performance across diverse domains, including \textit{face, autonomous driving, and image classification} tasks.
% We validate the efficiency and generalization of MaskDiME, which achieves over \textbf{30× faster} inference without additional training or fine-tuning, while maintaining superior performance across diverse domains, including \textit{face, autonomous driving, and image classification} tasks.

\section{Related Work}
\noindent \textbf{Explainable Artificial Intelligence} (XAI) has become a fundamental research direction aiming to make deep models transparent and trustworthy.
Existing XAI methods can be broadly divided into \textit{interpretable-by-design} approaches, which build explainability into model architectures~\cite{zhang2018interpretable,rudin2019stop,kapse2024si,zhu2025interpretable,singh2025protopatchnet,disciple2025learning}, and \textit{post-hoc} approaches, which analyze pretrained models. 
Post-hoc methods can be global, describing the overall behavior of a classifier, or local, explaining single predictions. 
Common local techniques include saliency maps~\cite{selvaraju2017grad,jalwana2021cameras,zheng2022shap,zhang2025towards,chen2025explainable,zhang2025finer}, 
concept attribution~\cite{kim2018interpretability,ghorbani2019towards,huy2025interactive,yu2025coe,grobrugge2025towards}, 
and model distillation~\cite{tan2018learning,haselhoff2021towards}. In our study, we focus on the emerging paradigm of Visual Counterfactual Explanations (VCEs)~\cite{wachter2017counterfactual,augustin2022diffusion}, which, unlike traditional XAI methods that mainly emphasize feature correlations, uncover the causal factors behind model decisions and present the visual cues a model relies on in a concise and human-intuitive manner.

% However, these methods typically capture statistical correlation rather than causal reasoning, limiting their ability to reveal which visual attributes truly drive a model’s decision.

\noindent \textbf{Visual Counterfactual Explanations} (VCEs) have been explored using various paradigms including VAEs~\cite{goodfellow2015explaining,haselhoff2024gaussian}, GANs~\cite{lang2021explaining,jacob2022steex,chowdhury2025looking}, and diffusion-based frameworks~\cite{jeanneret2022diffusion,jeanneret2023adversarial,weng2024fast,farid2023latent,jeanneret2024text,sobieski2024rethinking,yeganeh2025latent,kazimi2025explaining}. Diffusion models have become the dominant approach due to their superior generative fidelity and semantic control. 
DiME~\cite{jeanneret2022diffusion} was the first to apply diffusion models to counterfactual generation by globally guiding a pretrained model with classifier gradients, but its nested denoising and backpropagation at each step lead to a complexity of $O(T^2)$. ACE~\cite{jeanneret2023adversarial} adopts a two-stage framework that combines diffusion priors with adversarial optimization and refines local regions using pixel-difference masks, while RCSB~\cite{sobieski2024rethinking} performs region-constrained editing based on I2SB~\cite{liu20232}. Both enhance spatial controllability but struggle to handle semantic or non-rigid variations due to the use of fixed masks, and both incur high GPU memory usage. FastDiME~\cite{weng2024fast} achieves linear-time inference via Tweedie’s formula \cite{efron2011tweedie,song2020score} and uses pixel-difference masks to constrain modifications, yet it has limited precision due to coarse localization. %and offers no significant improvement over previous methods. 
Latent or text-conditioned variants (LDCE~\cite{farid2023latent}, TiME~\cite{jeanneret2024text}, LD~\cite{yeganeh2025latent}) inherit the Stable Diffusion framework~\cite{rombach2022high} but typically suffer from high computational cost or require additional tuning of text embeddings. Although MaskDiME is also based on a diffusion model and inherits the gradient-guided sampling paradigm of DiME, it operates in a single-stage manner with linear-time complexity ($O(T)$), overcoming the limitations of DiME’s global editing and nested denoising, ACE and RCSB’s fixed-region constraints, and FastDiME’s coarse pixel-difference masking, while maintaining lightweight efficiency and producing high-quality results.

\begin{figure*}[t]
  \centering
  \includegraphics[width=\linewidth]{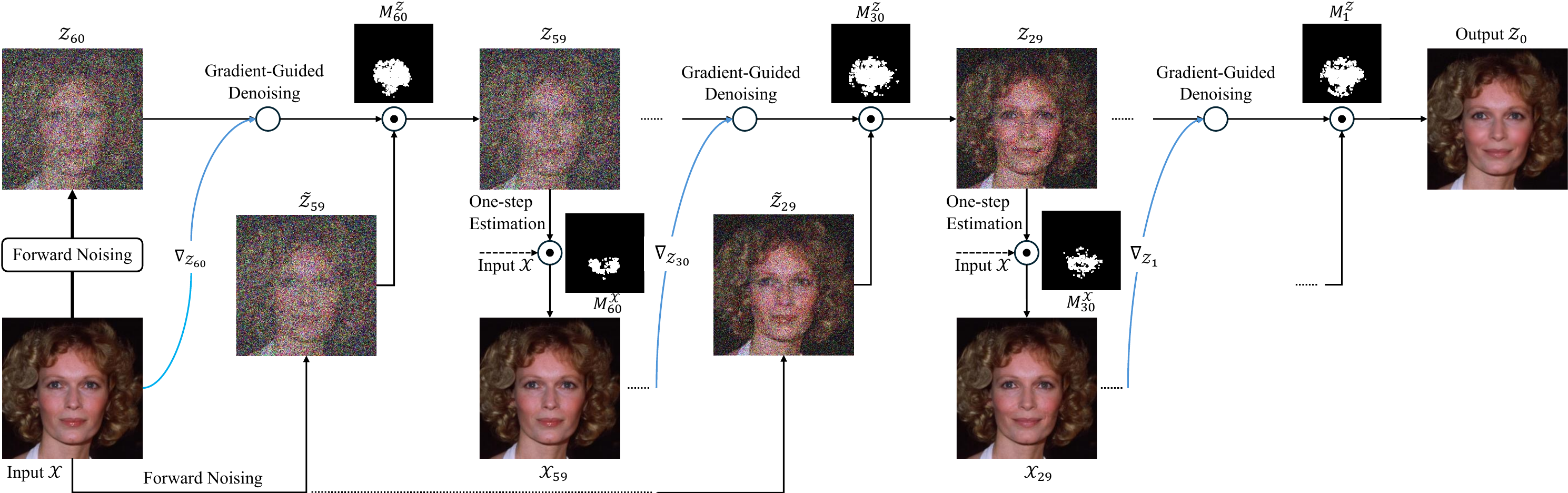}
\caption{
\textbf{Overview of MaskDiME.}
We illustrate a complete counterfactual generation process (\textit{No Smile → Smile}) with $\tau = 60$, meaning that the diffusion starts from $z_{60} = \tilde{z}_{60}$ and $x_{60} = x$, where $z_t$ and $x_t$ denote the noisy and clean images at step $t$, respectively
, and $\tilde{z}_t$ is obtained from the original forward diffusion process (\cref{eq:forward_diffusion})
. Each step applies Gradient-Guided Denoising, modeled as 
$\mathcal{N}\!\big(\mu_\theta(z_t) - \Sigma_\theta(z_t)\nabla_{z_t},\, \Sigma_\theta(z_t)\big)$, 
where the gradient $\nabla_{z_t}$ is derived from \cref{eq:gradients}. 
The adaptive masks $M_t^{x} \subseteq M_t^{z}$ are derived from classifier gradients on $x_t$,
where top-$k\%$ and top-$\rho k\%$ gradient regions construct $M_t^{z}$ and $M_t^{x}$, respectively ($\rho \in (0,1]$).
}

  \label{fig:fig1}
\end{figure*}

\section{Methodology}

\subsection{Preliminaries}

% We begin by introducing the fundamental generation process of diffusion models, which consists of two Markov chains: the \textit{forward diffusion} and the \textit{reverse denoising} processes.
We begin by introducing the gradient-guided diffusion process~\cite{dhariwal2021diffusion}, which consists of two Markov chains: the \textit{forward diffusion} and the \textit{gradient-guided reverse} processes. 
In the forward process, a clean image $x$ is progressively corrupted with Gaussian noise, resulting in a noisy sample $\tilde{z}_T$, where the tilde denotes a sample from the original forward diffusion process. The noisy sample at step $t$ is given by:
\begin{equation}
\tilde{z}_t = \sqrt{\bar{\alpha}_t}\, x + \sqrt{1 - \bar{\alpha}_t}\, \epsilon, 
\quad \epsilon \sim \mathcal{N}(0, I),
\label{eq:forward_diffusion}
\end{equation}
where $\bar{\alpha}_t = \prod_{s=1}^{t}(1 - \beta_s)$ is the cumulative product of the noise schedule, which controls the signal-to-noise ratio over time. By appropriately setting the schedule $\{\beta_t\}$, the final state $\tilde{z}_T$ can be made to approximate a standard Gaussian distribution $\mathcal{N}(0, I)$.

In the gradient-guided reverse process, the sampling dynamics are adjusted to encourage the generation of samples with desired attributes, iteratively denoising $\tilde{z}_T$ into $\{\tilde{z}_{T-1}, \dots, \tilde{z}_0\}$. A neural network (e.g., U-Net~\cite{ronneberger2015u}) with parameters $\theta$ is trained to estimate the mean and variance of conditional distribution at each timestep. Given a noisy input $\tilde{z}_t$, the model outputs the mean $\mu_\theta(\tilde{z}_t)$ and variance $\Sigma_\theta(\tilde{z}_t)$ that define the reverse sampling distribution:
\begin{equation}
\tilde{z}_{t-1} \sim \mathcal{N}\!\big(
\mu_\theta(\tilde{z}_t) 
- \Sigma_\theta(\tilde{z}_t)\nabla_{\tilde{z}_t} L(\tilde{z}_t; y),
\, \Sigma_\theta(\tilde{z}_t)
\big),
\label{eq:generalized-diff}
\end{equation}
where $L$ is a loss function defined on $\tilde{z}_t$ to encourage certain desired characteristics in the generated image, 
for example, conditioning the generation on a target label $y$.

% In the reverse process, the diffusion model aims to gradually denoise $z_T^{\text{orig}}$ to reconstruct the original image. A neural network (e.g., U-Net~\cite{ronneberger2015u}) with parameters $\theta$ is trained to estimate the mean and covariance of the conditional distribution at each timestep. Given a noisy input $z_t^{\text{orig}}$, the model predicts the mean $\mu_\theta(z_t^{\text{orig}})$ and the covariance $\Sigma_\theta(z_t^{\text{orig}})$, which define the reverse sampling distribution:
% $z_{t-1}^{\text{orig}} \sim \mathcal{N}(\mu_\theta(z_t^{\text{orig}}), \Sigma_\theta(z_t^{\text{orig}})).
% $ By recursively sampling $z_{t-1}^{\text{orig}}, z_{t-2}^{\text{orig}}, \ldots, z_0^{\text{orig}}$, the model ultimately generates a sample $z_0^{\text{orig}} \approx x$ that approximates the clean image. 

% In practice, to accelerate the recovery of clean images, some diffusion models (e.g., FastDiME~\cite{weng2024fast}) adopt a noise prediction formulation, 
% where the neural network  
% is trained to estimate the noise component $\epsilon_\theta(z_t^{\text{orig}})$ at each timestep $t$. 
% Following the Tweedie-based gradient approximation~\cite{efron2011tweedie,song2020score}, 
% the clean image estimate can be obtained in closed form as:
% \begin{equation}
% \hat{x}_0^{(t)} = \frac{z_t^{\text{orig}} - \sqrt{1 - \bar{\alpha}_t} \cdot \epsilon_\theta(z_t^{\text{orig}})}{\sqrt{\bar{\alpha}_t}}. \label{eq:tweedie}
% \end{equation}
% This formulation serves as the foundation for the guided denoising and spatial mask control mechanisms proposed in the following sections.

\subsection{MaskDiME}
Following the approach of DiME~\cite{jeanneret2022diffusion}, we treat counterfactual generation as an image editing task, as illustrated in \cref{fig:fig1}. 
Starting from a query image $x$, we first perform the forward diffusion process up to a predefined timestep $\tau$ (where $1 \leq \tau \leq T$) to obtain the corresponding noisy image $\tilde{z}_\tau$. 
Following the notation of DiME, we denote $z_t$ as the noisy image at timestep $t$, initialized with $\tilde{z}_\tau$ and progressively guided by gradients toward the clean counterfactual image $z_0$. 
Similarly, $x_t$ represents the estimated clean image derived from $z_t$, initialized with the input image $x$ and used for gradient
%$\nabla_{z_t}$ 
computation (see \cref{subsubsec:gradient} for details).
Throughout this process, the Adaptive Dual-mask Mechanism imposes spatial constraints on the updates by employing two binary masks at each timestep $t$, namely the noisy-level mask $M_t^{z}$ and the clean-level mask $M_t^{x}$ ($M_t^{x} \subseteq M_t^{z})$, ensuring that semantic modifications remain localized while preserving global structural consistency (see \cref{subsubsec:dualmask} for details).

Specifically, at each reverse diffusion step $t$, we update the noisy image $z_{t-1}$ as follows:
\begin{equation}
\begin{aligned}
z_{t-1} = &\; 
M_t^{z} \odot 
\mathcal{N}\!\big(
\mu_\theta(z_t) - \Sigma_\theta(z_t)\nabla_{z_t},\,
\Sigma_\theta(z_t)
\big) \\
&+ \big(1 - M_t^{z}\big) 
\odot \tilde{z}_{t-1} ,
\end{aligned}
\label{eq:reverse_step}
\end{equation}
where $\odot$ denotes element-wise multiplication, $\mu_\theta(z_t)$ and $\Sigma_\theta(z_t)$ are the predicted posterior mean and variance at step $t$, and $\tilde{z}_{t-1}$ is obtained by adding Gaussian noise to $x$ through the forward process in \cref{eq:forward_diffusion}.
This operation enables spatially controlled denoising, where updates are applied only within the masked region while the unmasked area preserves the original diffusion trajectory.

Once \(z_{t-1}\) is obtained, a one-step estimation based on Tweedie’s formula \cite{efron2011tweedie,song2020score} is applied to estimate the clean image at step $t\!-\!1$:
\begin{equation}
\hat{x}_0^{(t-1)} = \frac{z_{t-1} - \sqrt{1 - \bar{\alpha}_{t-1}} \epsilon_\theta(z_{t-1})}{\sqrt{\bar{\alpha}_{t-1}}}, \label{eq:get_clean}
\end{equation}
where $\epsilon_\theta(z_{t-1})$ denotes the noise predicted by the denoising network. Finally, we update \(x_{t-1}\) for the next timestep using the clean-level mask \(M_t^{x}\):
\begin{equation}
x_{t-1} = M_t^{x} \odot \hat{x}_0^{(t-1)} + (1 - M_t^{x}) \odot x.
\end{equation}
% This blending maintains the surrounding context while allowing targeted semantic modifications within the guided regions.
This blending maintains surrounding context while allowing targeted semantic changes within the guided regions.

\subsubsection{Gradient-Based Guidance}
\label{subsubsec:gradient}
We adopt the joint loss function from DiME, which consists of a classification loss $L_{\text{class}}$, a perceptual loss $L_{\text{perc}}$, and an $L_1$ loss~\cite{jeanneret2022diffusion}. The classification loss encourages the generated image to move toward the semantics of the target class, while the perceptual loss enforces similarity to the original image $x$ in terms of structure or appearance. In addition, the $L_1$ loss provides pixel-wise supervision to stabilize low-level differences and reduce artifacts during sampling. The joint loss is
\begin{equation}
\begin{split}
{L}(x_t; y, x) =\ & \lambda_c L_{\text{class}}(C(y|x_t)) + \lambda_p L_{\text{perc}}(x_t, x) \\
& + \lambda_l L_{\text{L1}}(x_t, x),
\end{split}
\end{equation}
where $\lambda_c$, $\lambda_p$, and $\lambda_l$ are hyperparameters controlling the relative weights of the classification, perceptual, and $L_1$ terms, respectively. Here, $C(y|x_t)$ denotes the classifier’s predicted probability for the target class $y$ given input $x_t$.

To enable gradient-based guidance in the reverse diffusion process, we require the gradient of the joint loss ${L}$ with respect to the noisy input $z_t$. 
% Since ${L}$ is defined on the clean image estimate $x_t$, we approximate $x_t$ using the predicted reconstruction $\hat{x}_0^{(t)}$ given by Eq.~(3). Noting that $z_t \approx z_t^{\text{orig}}$, we have:
% \begin{equation}
% x_t \approx \hat{x}_0^{(t)}.
% \end{equation}
We follow the reparameterization trick used in DiME to obtain the gradient of the loss with respect to \( z_t \) as:
\begin{equation}
\nabla z_t = s  \frac{1}{\sqrt{\bar{\alpha}_t}} \nabla_{x_t} {L}(x_t; y, x). \label{eq:gradients}
\end{equation}
To simplify the hyperparameters, we retain DiME’s original weights $\lambda_c \in \{8, 10, 15\}$ to iteratively find the counterfactuals, $\lambda_p = 30$, and $\lambda_l = 0.05$, and introduce a single scaling factor $s$ that controls the overall guidance strength.

\subsubsection{Adaptive Dual-mask Mechanism}
\label{subsubsec:dualmask}
To realize the spatial constraints described above, we construct two masks at each diffusion timestep, $
M_t^{x} \subseteq M_t^{z}
$. $M_t^{z}$ constrains the update from $z_t$ to $z_{t-1}$ and $M_t^{x}$ constrains $z_{t-1}$ to $x_{t-1}$. Both are constructed based on the classifier gradients of the reconstructed image $x_t$, rather than pixel-level differences as~\cite{weng2024fast}.
Specifically, we first extract the spatial gradient magnitude derived from the classification loss:
\begin{equation}
\nabla_{z_t}^{\text{class}} = \frac{1}{\sqrt{\bar{\alpha}_t}} \nabla_{x_t} L_{\text{class}}(C(y|x_t)).
\end{equation}
Then, the scalar gradient map is obtained by taking the element-wise absolute value and averaging over the channel dimension \cite{simonyan2013deep}:
\begin{equation}
G_t = \left| \nabla_{z_t}^{\text{class}} \right|_{\text{avg}} \in \mathbb{R}^{1 \times H \times W}.
\end{equation}

Unlike attribution-based methods, such as Integrated Gradients (IG)~\cite{sundararajan2017axiomatic}, used in RCSB~\cite{sobieski2024rethinking}, which estimate pixel importance by integrating gradients along multiple interpolation paths—requiring dozens of forward and backward passes and incurring substantial computational costs—our method constructs the spatial mask in a single step. This enables efficient identification of decision-relevant regions, allowing the generation of the mask at each sampling step. The mask \( M_t^{z} \) is defined pixel-wise: for each location \((i, j)\), we set \( M_t^{z}(i, j) = 1 \) if the gradient magnitude \( G_t(i, j) \) falls within the top-\( k\% \) of all gradient magnitudes in \( G_t \); otherwise, \( M_t^{z}(i, j) = 0 \).

Our experiments reveal that using the same mask for both noisy and clean images, the classifier gradient with respect to $x_t$ gradually diminishes as $t$ decreases. As denoising progresses, $x_t$ moves closer to the counterfactual class, leading to weaker classifier gradients, with the perceptual and reconstruction losses ($L_{\text{perc}}$ and $L_1$) dominating in \cref{eq:gradients}.
% We attribute this to the fact that $x_t$ becomes increasingly close to the counterfactual class as the denoising process proceeds, causing the gradients from the perceptual loss $L_{\text{perc}}$ and reconstruction loss $L_1$ to dominate in \cref{eq:gradients}.
% We attribute this to the fact that, as the denoising process proceeds, $x_t$ becomes increasingly close to the counterfactual class, resulting in weaker classifier gradients. Consequently, the gradients from the perceptual loss $L_{\text{perc}}$ and reconstruction loss $L_1$ dominate in \cref{eq:gradients}.
To mitigate this, we define the mask $M_t^{x}$ by retaining only a fraction $\rho$ of the pixels in $M_t^{z}$ that correspond to the highest gradient values. In other words, $M_t^{x}$ selects the top-$\rho k\%$ of all gradient magnitudes, where $\rho \in (0,1]$ controls the shrinkage rate (e.g., $\rho = 0.5$ keeps the pixels associated with the top 50\% largest gradient values within $M_t^{z}$). To improve spatial coherence, we apply morphological dilation with a kernel size of 5 to both $M_t^{z}$ and $M_t^{x}$.

% To mitigate this effect, we further define a clean-level mask $M_t^{\text{c}}$ by retaining only a fraction $\rho$ of the most salient pixels within $M_t^{\text{noisy}}$:
% \begin{equation}
% M_t^{\text{clean}}(i,j) =
% \begin{cases}
% 1, & \text{if } G_t(i,j) \text{ is in top}\rho k\%, \\
% 0, & \text{otherwise},
% \end{cases}
% \end{equation}
% where ($M_t^{\text{clean}} \subseteq M_t^{\text{noisy}}$), and $\rho \in (0,1]$ controls the shrinkage rate (e.g., $\rho=0.5$ retains the strongest 50\% of pixels in $M_t^{\text{noisy}}$).

% To improve spatial coherence, both $M_t^{\text{noisy}}$ and $M_t^{\text{clean}}$ are smoothed using Gaussian filtering and then re-binarized via thresholding. Specifically, Both $M_t^{\text{noisy}}$ and $M_t^{\text{clean}}$ are updated by applying a Gaussian blur 
% with a default kernel size of 5 and re-binarizing the result, i.e.,
% $
%     M_t^{(\cdot)} = \mathbb{I}\big[\text{blur}(M_t^{(\cdot)}) > 0\big],
%     \quad (\cdot) \in \{\text{noisy}, \text{clean}\}.
% $
 % These spatial masks are then used in the reverse diffusion loop to constrain counterfactual edits to semantically relevant regions.
\begin{figure}[t]
  \centering
   \includegraphics[width=1\linewidth]{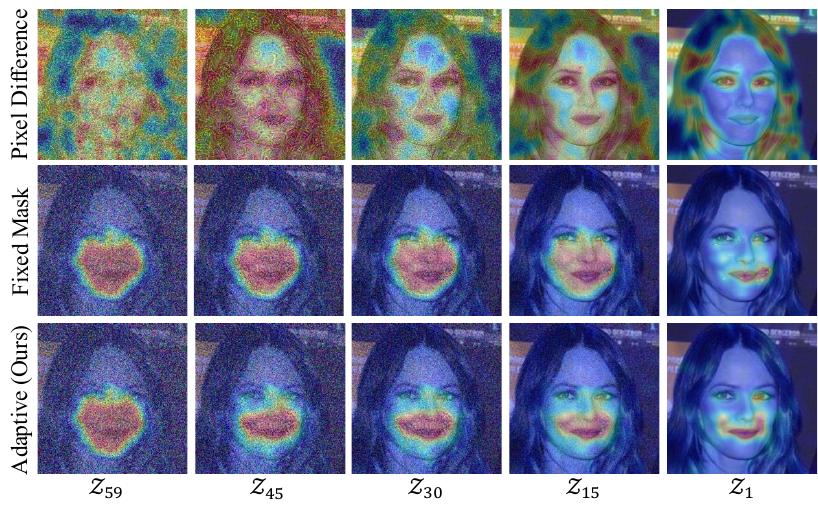}

\caption{
Heatmap visualization of diffusion trajectories with different masking strategies. 
Each column shows noisy samples $z_t$ at different timesteps, and each row corresponds to a masking method. 
The heatmap represents the per-pixel update magnitude during each reverse step (red indicates stronger updates, blue indicates weaker ones). 
Our Adaptive Dual-mask yields the most focused and semantically consistent updates across diffusion steps.
}

   \label{fig:heatmaps}
\end{figure}

\subsubsection{Visualization Study}
To further validate our hypothesis regarding the interaction between classifier guidance and diffusion dynamics, we visualize the evolution of noisy samples $z_t$ under different masking strategies, including the pixel-difference mask, fixed mask, and our proposed Adaptive Dual-mask, as shown in \cref{fig:heatmaps}. 
As illustrated, the pixel-difference mask yields dispersed updates across the image, which limits the ability of guidance to focus on decision-relevant regions. Its behavior is nearly identical to the no-mask case, so we omit the no-mask results for brevity. The fixed mask mitigates this issue by constraining updates within a predefined area, but its static nature prevents adaptation to evolving semantics, resulting in either overly broad updates or excessive focus on local regions (e.g., abnormal modification near the mouth corner in the rightmost column). 

In contrast, our Adaptive Dual-mask adjusts its spatial focus at each diffusion step based on classifier gradients, guiding the model toward decision-relevant regions. The per-timestep heatmaps show that updates progressively concentrate on the mouth and other areas related to expression, producing spatially coherent counterfactual trajectories with fewer irrelevant perturbations. These results demonstrate that the proposed gradient-driven masking enables diffusion models to focus on causal regions, yielding more precise and semantically consistent counterfactuals.

\begin{table*}[t]
\centering
\caption{\textbf{CelebA and CelebA-HQ assessment.} We report results extracted from DiVE~\cite{rodriguez2021beyond}, STEEX~\cite{jacob2022steex}, DiME~\cite{jeanneret2022diffusion}, 
ACE \cite{jeanneret2023adversarial}, FastDiME \cite{weng2024fast}, 
LDCE \cite{farid2023latent}, TiME \cite{jeanneret2024text}, 
and RCSB \cite{sobieski2024rethinking}.
Best and second-best results are shown in \textbf{bold} and \underline{underline}, respectively.
% Arrows $(\downarrow/\uparrow)$ indicate whether lower or higher values are better. 
}
\label{tab:celeba_results}

{\fontsize{9pt}{10pt}\selectfont
\setlength{\tabcolsep}{1.75pt} 
\begin{tabular}{@{}l@{}*{8}{c}|*{8}{c}}
\toprule
\multirow{2}{*}{Method} & \multicolumn{8}{c}{\textbf{Smile}} & \multicolumn{8}{c}{\textbf{Age}} \\
\cmidrule(lr){2-9} \cmidrule(lr){10-17}
& FID↓ & sFID↓ & FVA↑ & FS↑ & MNAC↓ & CD↓ & COUT↑ & FR(\%)↑
& FID↓ & sFID↓ & FVA↑ & FS↑ & MNAC↓ & CD↓ & COUT↑ & FR(\%)↑ \\
\midrule
&\multicolumn{16}{c}{\textbf{CelebA}} \\
\midrule
DiVE        & 29.4 & - & 97.3 & - & - & - & - & - & 33.8 & - & 98.2 & - & 4.58 & - & - & - \\
DiVE$^{100}$& 36.8 & - & 73.4 & - & 4.63 & 2.34 & - & - & 39.9 & - & 52.2 & - & 4.27 & - & - & - \\
STEEX       & 10.2 & - & 96.9 & - & 4.11 & - & - & - & 11.8 & - & 97.5 & - & 3.44 & - & - & - \\
DiME        & 3.17 & 4.89 & 98.3 & 0.73 & 3.72 & 2.30 & 0.53 & 97.2 & 4.15 & 5.89 & 95.3 & 0.67 & 3.13 & 3.27 & 0.44 & 99.0 \\
ACE $\ell_1$& \underline{1.27} & \underline{3.97} & \underline{99.9} & \underline{0.87} & 2.94 & \underline{1.73} & \underline{0.78} & 97.6
            & \underline{1.45} & \underline{4.12} & 99.6 & 0.78 & 3.20 & \underline{2.94} & 0.72 & 96.2 \\
ACE $\ell_2$& 1.90 & 4.56 & \underline{99.9} & 0.87 & \underline{2.77} & \textbf{1.56} & 0.62 & 84.3
            & 2.08 & 4.62 & 99.6 & 0.80 & 2.94 & \textbf{2.82} & 0.56 & 77.5 \\
FastDiME    & 4.18 & 6.13 & 99.8 & 0.76 & 3.12 & 1.91 & 0.45 & 99.0 
            & 4.82 & 6.76 & 99.2 & 0.74 & 2.65 & 3.80 & 0.36 & 98.6 \\
FastDiME-2  & 3.33 & 5.49 & \underline{99.9} & 0.77 & 3.06 & 1.89 & 0.44 & 99.4 
            & 4.04 & 6.01 & 99.6 & 0.75 & 2.63 & 3.80 & 0.37 & \underline{99.3} \\
FastDiME-2+ & 3.24 & 5.23 & \underline{99.9} & 0.79 & 2.91 & 2.02 & 0.41 & 98.9 
            & 3.60 & 5.59 & 99.7 & 0.77 & 2.44 & 3.76 & 0.32 & 98.7 \\
RCSB        & 2.98 & 4.79 & \textbf{100.0} & \textbf{0.91} & \textbf{2.24} & 2.78 & \textbf{0.87} & \underline{99.8}  
            & 2.94 & 4.94 & \underline{99.9} & \textbf{0.88} & \textbf{2.14} & 3.63 & \textbf{0.81} & 99.3 \\
MaskDiME (Ours)   & \textbf{0.71} & \textbf{3.29} & \textbf{100.0} & \textbf{0.91} & 2.78 & 2.41 & \textbf{0.87} & \textbf{100.0}
            & \textbf{0.77} & \textbf{3.33} & \textbf{100.0} & \underline{0.83} & \underline{2.22} & 3.21 & \underline{0.79} & \textbf{100.0} \\ 
\midrule
&\multicolumn{16}{c}{\textbf{CelebA-HQ}} \\
\midrule
DiVE & 107.0 & - & 35.7 & - & 7.41 & - & - & - & 107.5 & - & 32.3 & - & 6.76 & - & - & - \\
STEEX & 21.9 & - & 97.6 & - & 5.27 & - & - & - & 26.8 & - & - & - & 5.63 & - & - & - \\
DiME & 18.10 & 27.7 & 96.7 & 0.67 & 2.63 & \underline{1.82} & 0.65 & 97.0 & 18.7 & 27.8 & 95.0 & 0.66 & 2.10 & 4.29 & 0.56 & 97.0 \\
ACE $\ell_1$ & 3.21 & 20.2 & \textbf{100.0} & 0.89 & 1.56 & 2.61 & 0.55 & 95.0 & 5.31 & \underline{21.7} & \underline{99.6} & 0.81 & \underline{1.53} & 5.40 & 0.40 & 95.0 \\
ACE $\ell_2$ & 6.93 & 22.0 & \textbf{100.0} & 0.84 & 1.87 & 2.21 & 0.60 & 95.0 & 16.4 & 28.2 & \underline{99.6} & 0.77 & 1.92 & \underline{4.21} & 0.53 & 95.0 \\
LDCE-txt & 13.6 & 25.8 & \underline{99.1} & 0.76 & 2.44 & \textbf{1.68} & 0.34 & - & 14.2 & 25.6 & 98.0 & 0.73 & 2.12 & \textbf{4.02} & 0.33 & - \\
TiME  & 10.98 & 23.8 & 96.6 & 0.79 & 2.97 & 2.32 & 0.63 & 97.1 & 20.9 & 32.9 & 79.3 & 0.63 & 4.19 & 4.29 & 0.31 & 89.9 \\
RCSB  & \underline{3.04} & \underline{20.0} & \textbf{100.0} & \underline{0.93} & \textbf{1.22} & 3.22 & \textbf{0.83} & \underline{98.9}
      & \underline{4.92} & 27.3 & \textbf{100.0} & \textbf{0.96} & \textbf{1.47} & 5.16 & \textbf{0.80} & \underline{99.4} \\
MaskDiME (Ours)  & \textbf{2.51} & \textbf{18.1} & \textbf{100.0} & \textbf{0.94} & \underline{1.41} & 2.67 & \underline{0.69} & \textbf{99.4}
          & \textbf{4.43} & \textbf{19.4} & \textbf{100.0} & \underline{0.88} & 1.82 & 4.67 & \underline{0.63} & \textbf{99.6} \\
\bottomrule
\end{tabular}
}
\end{table*}

\begin{table}[ht]
\centering
\caption{\textbf{BDD assessment}. We extract results from STEEX~\cite{jacob2022steex}, ACE \cite{jeanneret2023adversarial} 
and TiME \cite{jeanneret2024text}, where results marked with $^\dagger$ are directly reported from ACE.
Best and second-best results are shown in \textbf{bold} and \underline{underline}, respectively. 
 % Arrows $(\downarrow/\uparrow)$ indicate whether lower or higher values are better.
}
\label{tab:bdd_comparison}

{\fontsize{9pt}{10pt}\selectfont 
\setlength{\tabcolsep}{4.4pt}      

\begin{tabular}{lccccc}
\toprule
Method & FID↓  & sFID↓ & $S^3$↑ & COUT↑ & FR(\%)↑ \\
\midrule
\multicolumn{6}{c}{\textbf{BDD100K}} \\
\midrule
STEEX            & 58.80 & --    & --     & --     & 99.5 \\
DiME$^\dagger$             & 7.94  & 11.40 & 0.95 & 0.24 & 90.5 \\
ACE $l_1$        &\textbf{1.02}  & \textbf{6.25}  & \textbf{1.0} & 0.75 & 99.9 \\
ACE $l_2$        & \underline{1.56}  & \underline{6.53}  & \textbf{1.0} & 0.79 & 99.9 \\
TiME             & 51.5  & 76.18 & 0.77 & 0.15 & 81.8 \\
MaskDiME (Ours)        & 3.19 & 7.85 & \underline{0.99} & \textbf{0.85} & \textbf{100.0} \\
\midrule
\multicolumn{6}{c}{\textbf{BDD-OIA}} \\
\midrule
DiME$^\dagger$             & 13.70 & 26.06 & 0.93 & 0.32 & 91.7 \\
ACE $l_1$        & \textbf{2.09}  & \textbf{22.13} & \textbf{1.0} & 0.74 & \underline{99.9} \\
ACE $l_2$        & \underline{3.30}  & \underline{22.75} & \textbf{1.0} & \underline{0.78} & \textbf{100.0} \\
MaskDiME (Ours)       & 5.43 & 23.38 & \underline{0.99} & \textbf{0.80} & \textbf{100.0} \\
\bottomrule
\end{tabular}
}
\end{table}

\begin{table*}[htbp]
\centering
\caption{\textbf{ImageNet Assessement}. We extract results from ACE~\cite{jeanneret2023adversarial}, LDCE~\cite{farid2023latent}, and RCSB~\cite{sobieski2024rethinking}, where results marked with $^\dagger$ (e.g. DVCE~\cite{augustin2022diffusion}) are directly reported from RCSB. Best and second-best results are shown in \textbf{bold} and \underline{underline}, respectively. 
% Arrows $(\downarrow/\uparrow)$ indicate whether lower or higher values are better.
}
\label{tab:imagenetresults_wide}

{\fontsize{9pt}{10pt}\selectfont
\setlength{\tabcolsep}{2.4pt}
\begin{tabular}{l|ccccc|ccccc|ccccc}
\toprule
\multirow{2}{*}{\textbf{Method}} &
\multicolumn{5}{c|}{\textbf{Sorrel -- Zebra}} &
\multicolumn{5}{c|}{\textbf{Persian Cat -- Egyptian Cat}} &
\multicolumn{5}{c}{\textbf{Cougar -- Cheetah}} \\
\cmidrule(lr){2-6} \cmidrule(lr){7-11} \cmidrule(lr){12-16}
& FID↓ & sFID↓ & $S^3$↑ & COUT↑ & FR(\%)↑
& FID↓ & sFID↓ & $S^3$↑ & COUT↑ & FR(\%)↑
& FID↓ & sFID↓ & $S^3$↑ & COUT↑ & FR(\%)↑
\\
\midrule
DiME$^\dagger$ & 222.9 & 243.2 & 0.19 & -0.31 & 0.0 & 322.8 & 352.1 & 0.44 & -0.05 & 0.0 & 268.2 & 292.0 & 0.11 & -0.16 & 0.0 \\
ACE $l_1$ & 84.5 & 122.7 & \underline{0.92} & -0.45 & 47.0 & 93.6 & 156.7 & \underline{0.85} & 0.25 & 85.0 & 70.2 & 100.5 & 0.91 & 0.02 & 77.0 \\
ACE $l_2$ & 67.7 & 98.4 & 0.90 & -0.25 & 81.0 & 107.3 & 160.4 & 0.78 & 0.34 & 97.0 & 74.1 & 102.5 & 0.88 & 0.12 & 95.0 \\
LDCE-cls & 84.2 & 107.2 & 0.78 & -0.06 & 88.0 & 102.7 & 140.7 & 0.63 & 0.52 & \underline{99.0} & 71.0 & 91.8 & 0.62 & \underline{0.51} & \textbf{100.0} \\
LDCE-txt & 82.4 & 107.2 & 0.71 & -0.21 & 81.0 & 121.7 & 162.4 & 0.61 & 0.56 & \underline{99.0} & 91.2 & 117.0 & 0.59 & 0.34 & 98.0 \\
FastDiME$^\dagger$ & 96.5 & 103.5 & 0.22 & -0.44 & 14.0 & 193.6 & 207.1 & 0.10 & 0.01 & 20.0 & 133.0 & 141.1 & 0.12 & -0.11 & 18.0 \\
DVCE$^\dagger$ & 33.1 & 43.9 & 0.62 & -0.21 & 57.8 & 46.6 & 59.2 & 0.59 & 0.60 & 98.5 & 46.9 & 54.1 & 0.70 & 0.49 & \underline{99.0} \\
RCSB & \textbf{8.0} & \textbf{16.2} & 0.88 & \underline{0.74} & \underline{94.7} & \textbf{23.0} & \textbf{40.0} & \textbf{0.87} & \textbf{0.92} & \textbf{100.0} & \textbf{17.2} & \textbf{26.6} & \underline{0.92} & \textbf{0.92} & \textbf{100.0} \\
MaskDiME (Ours) & \underline{32.5} & \underline{39.7} & \textbf{0.94} & \textbf{0.85} & \textbf{99.7} & \underline{36.2} & \underline{50.1} & \textbf{0.87} & \underline{0.89} & \textbf{100.0} & \underline{37.4} & \underline{46.3} & \textbf{0.93} & \textbf{0.92} & \textbf{100.0} \\
\bottomrule
\end{tabular}
}
\end{table*}

\section{Experimentation}
\subsection{Evaluation Protocols and Datasets}
\noindent \textbf{Datasets.} Following the experimental protocol of ACE~\cite{jeanneret2023adversarial}, we evaluate the proposed MaskDiME on five representative datasets covering diverse visual domains.
CelebA~\cite{liu2015deep} is a widely used facial attribute dataset containing diverse face images with rich attribute annotations. We use its aligned $128\times128$ version, focusing on the \textit{smile} and \textit{age} attributes for binary counterfactual generation. 
CelebA-HQ~\cite{lee2020maskgan} provides higher-quality $256\times256$ face images for evaluating counterfactual generation on high-resolution data. BDD100K~\cite{yu2020bdd100k} contains $512\times256$ street-scene images for autonomous driving, where we evaluate binary driving decisions of \textit{forward} versus \textit{slow down}. BDD-OIA~\cite{xu2020explainable} extends BDD100K with additional contextual annotations for driving behaviors, providing a stronger benchmark for realism and actionability. Finally, we follow prior work~\cite{jeanneret2023adversarial,farid2023latent,sobieski2024rethinking} on ImageNet~\cite{deng2009imagenet}, where images are resized to $256\times256$ and visually similar category pairs (e.g., \textit{sorrel vs. zebra}, \textit{persian cat vs. egyptian cat}, \textit{cougar vs. cheetah}) are selected to evaluate the scalability and semantic precision of MaskDiME in multi-class scenarios.\\
\noindent \textbf{Evaluation Metrics.} To comprehensively assess the quality of counterfactual explanations, we follow the evaluation protocol of ACE~\cite{jeanneret2023adversarial} and incorporate extensions from subsequent works~\cite{farid2023latent,weng2024fast,jeanneret2024text,sobieski2024rethinking}, covering validity, realism, similarity, sparsity, decision consistency, and diversity. Specifically, the Flip Rate (\underline{FR}) measures validity, i.e., whether a counterfactual flips the classifier’s prediction. For realism, we report both the standard Fréchet Inception Distance (\underline{FID})~\cite{heusel2017gans} and the split-FID (\underline{sFID}) proposed in ACE, where the dataset is randomly divided into two disjoint subsets; counterfactuals are generated on one half and compared against the other, and repeated ten times with the mean score reported, which alleviates the bias introduced by unchanged pixels. On face datasets (CelebA/CelebA-HQ), identity preservation and proximity are measured using Face Verification Accuracy (\underline{FVA})~\cite{cao2018vggface2} and Face Similarity (\underline{FS}), where FVA evaluates verification accuracy with a face recognition network, and FS computes the cosine similarity of feature embeddings between original–counterfactual pairs. On non-face datasets (BDD100K/BDD-IOA/ImageNet), we employ SimSiam Similarity (\underline{S$^3$})~\cite{chen2021exploring} to quantify representation-level proximity. For sparsity, we report the Mean Number of Attributes Changed (\underline{MNAC})~\cite{rodriguez2021beyond}, which measures the average number of modified attributes, and complement it with the Counterfactual Difference (\underline{CD})~\cite{jeanneret2022diffusion}, which assesses whether modifications stay consistent with training-data correlations, ensuring semantic plausibility. For decision consistency, we use Counterfactual Output Transition (\underline{COUT})~\cite{khorram2022cycle}, which measures the shift in the classifier’s output distribution between original and counterfactual inputs. Diversity is quantified as the average pairwise LPIPS~\cite{zhang2018unreasonable} distance (\underline{$\sigma_L$}) across counterfactual samples from five independent runs with different seeds, where a higher value indicates greater perceptual diversity.\\
\noindent \textbf{Implementation Details.} 
MaskDiME is a \emph{training-free} extension of DiME: we directly reuse the unconditional DDPM~\cite{ho2020denoising} weights and target classifier weights from prior works without any additional training or fine-tuning (see Tab.~6 in the supplementary material).
Unless otherwise specified, all hyperparameters are inherited from DiME, using 200 sampling steps with the reverse process starting from timestep $\tau = 60$.
The newly introduced parameter $s$ is empirically tuned through qualitative visualization, set to 8 for CelebA, 10 for CelebA-HQ, 14 for BDD100K/BDD-OIA, and 6.5 for ImageNet. For spatial masking, we set the top-$k$ percentage to $k = 0.05$ (5\%) for the \textit{smile} attribute on CelebA and CelebA-HQ, while $k = 0.1$ (10\%) is used for the \textit{age} attribute and all other datasets. The shrinkage factor $\rho$ is set to 0.25 on CelebA-HQ and 0.5 for all other datasets. More details can be
found in the supplemental materials.

\subsection{Results and Analysis}

\paragraph{Comparison with the State-of-the-Art}
We first conduct a quantitative comparison between MaskDiME and existing counterfactual generation methods. \cref{tab:celeba_results,tab:bdd_comparison,tab:imagenetresults_wide} demonstrate that
MaskDiME consistently achieves either superior or near-optimal performance in terms of realism (FID/sFID), validity (FR), similarity (FVA, FS, $S^3$), and decision consistency (COUT) across all datasets, while maintaining competitive results in sparsity (MNAC) and semantic plausibility (CD) on facial datasets.

On CelebA/CelebA-HQ facial datasets, MaskDiME achieves the lowest (best) FID/sFID and the highest (best) FR (100\% on CelebA), while achieving the best or second-best results in FVA, FS, and COUT, demonstrating excellent realism, validity, and decision consistency. For MNAC, MaskDiME is not the best but remains close to the optimal, maintaining compact editing regions. In terms of the CD, MaskDiME shows slightly higher values than DiME and ACE, which may be attributed to its local constraint mechanism; however, it still achieves clearly better scores than RCSB under the same locally constrained editing scheme, reflecting a balanced trade-off that produces more natural and semantically plausible counterfactuals.

Results on autonomous driving in \cref{tab:bdd_comparison} demonstrate that MaskDiME achieves 100\% FR on both BDD100K and BDD-OIA, indicating strong validity and stability. Although ACE attains slightly better FID/sFID, MaskDiME substantially improves over DiME and TiME, while maintaining top-tier performance in COUT and $S^3$, demonstrating its robustness in complex driving scenarios.

\begin{figure}[ht]
  \centering
   \includegraphics[width=1\linewidth]{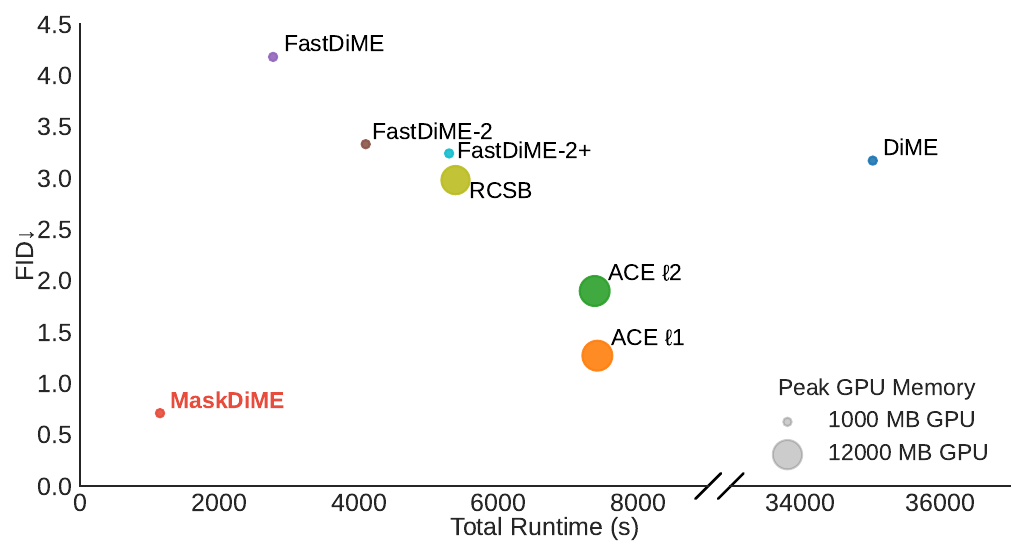}

\caption{Comparison of methods on the CelebA \textit{smile} attribute by FID (from \cref{tab:celeba_results}), runtime (batch size = 5). The area of the circles indicates the peak GPU memory allocated during the sampling process. MaskDiME is significantly faster than previous methods, while also achieving the lowest FID, and sustaining low GPU usage—approximately one-tenth of that required by ACE and RCSB. See Supplementary Tab.~7 for quantitative results.
}
   \label{fig:efficiency}
\end{figure}

\begin{figure*}[ht]
  \centering
  \includegraphics[width=1\linewidth]{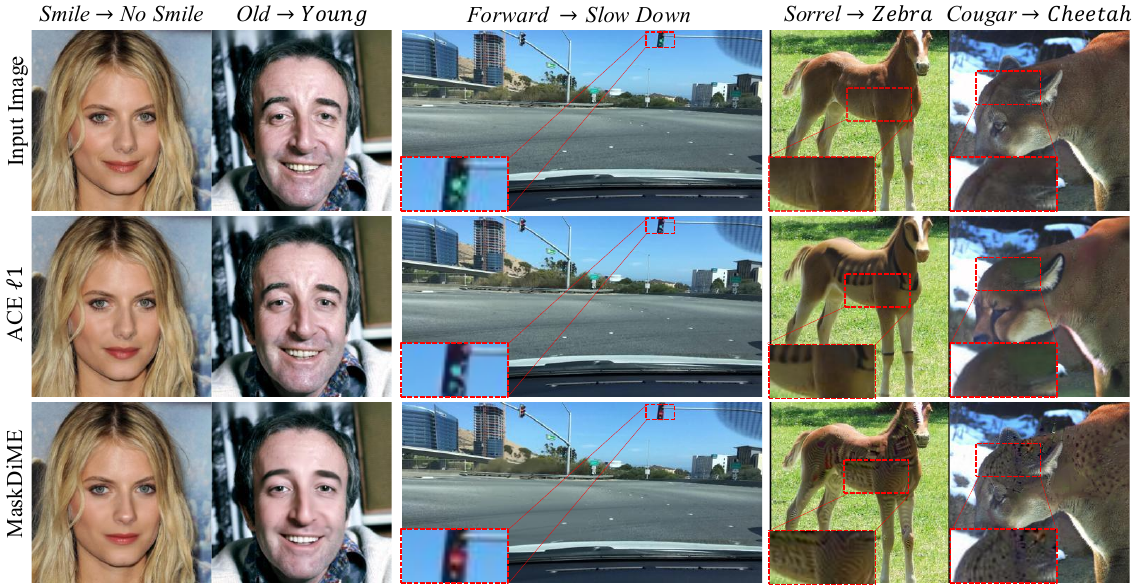}
  \caption{\textbf{Qualitative results.} Compared with ACE $l_1$, MaskDiME effectively preserves the overall image structure and produces more pronounced counterfactual explanations, with superior performance in semantic consistency, visual realism, and modification precision.}
  \label{fig:Qualitative}
\end{figure*}
On the more challenging ImageNet dataset in \cref{tab:imagenetresults_wide}, the difficulty of counterfactual generation increases significantly due to the large semantic gaps between categories and the strong contextual dependencies of model predictions. Notably, even the ACE paper did not report results for DiME, and the experiments published in RCSB show that both DiME and FastDiME almost completely fail on this dataset (FR close to 0 and extremely high FID), highlighting the challenge of maintaining generation stability and semantic consistency in complex multi-class scenarios. Despite this, MaskDiME demonstrates consistent and robust performance across all category pairs.
It generally ranks within the top two across all metrics, achieving the best results in both $S^3$ and FR, and obtaining the highest COUT scores on two category pairs.
Although MaskDiME yields higher FID and sFID scores than RCSB (while consistently ranking second), this is largely attributed to the inherent difficulty of multi-class counterfactual generation on ImageNet. These results indicate that MaskDiME maintains the effectiveness and causal consistency of counterfactual generation even under complex semantic conditions.

Qualitatively, \cref{fig:Qualitative} presents generated examples that cover all visual domains.  
We include ACE for comparison, as it is, to the best of our knowledge, the only method that reports results across all datasets with consistently strong performance.  
Overall, MaskDiME produces more realistic and semantically accurate counterfactuals than ACE—for instance, it can more effectively modify traffic light colors in the driving domain to yield valid and interpretable outcomes. See supplementary for more qualitative results.

\noindent \textbf{Computational Efficiency Analysis}
\label{sec:efficiency}
We compare the generation efficiency of DiME, ACE, FastDiME, and RCSB in \cref{fig:efficiency}.
Following FastDiME, we use a random subset of 1000 CelebA images and generate counterfactuals for the \textit{smile} attribute 
on an NVIDIA Tesla V100 (32~GB) GPU. The scatter plot reports the total runtime and peak GPU memory, with FID from \cref{tab:celeba_results}. The peak memory usage is measured as the maximum allocated memory across all sampling steps, which more accurately reflects the model’s actual memory footprint 
compared to the overall GPU utilization. MaskDiME eliminates the recursive sampling process of DiME by employing an efficient one-step estimation strategy.
Leveraging masked sampling and the gradient scaling factor $s$ , it enables faster counterfactual generation with fewer adjustments of the classifier weight $\lambda_c$.
Consequently, MaskDiME achieves over 30× and 2.5× speedups compared with DiME and FastDiME, respectively, while attaining superior FID. In contrast, ACE and RCSB demand higher GPU memory and exhibit slower generation.

\noindent \textbf{Diversity Assessment}
We evaluate diversity on the same CelebA subset as before by varying only the random noise during generation. Quantitatively, DiME achieves a diversity score of $\sigma_L = 0.2139$, while ACE $l_1$ attains $\sigma_L = 0.0174$. Although MaskDiME reaches $\sigma_L = 0.0395$, which is far below DiME, it is still higher than ACE $l_1$. This outcome is expected, as DiME lacks spatial constraints and often modifies irrelevant background regions, which leads to an overestimation of diversity. In contrast, MaskDiME limits edits to decision-relevant regions, resulting in a lower but more faithful estimate of semantic diversity, as it avoids spurious variations in irrelevant areas.

\begin{figure}[t]
  \centering
   \includegraphics[width=1\linewidth]{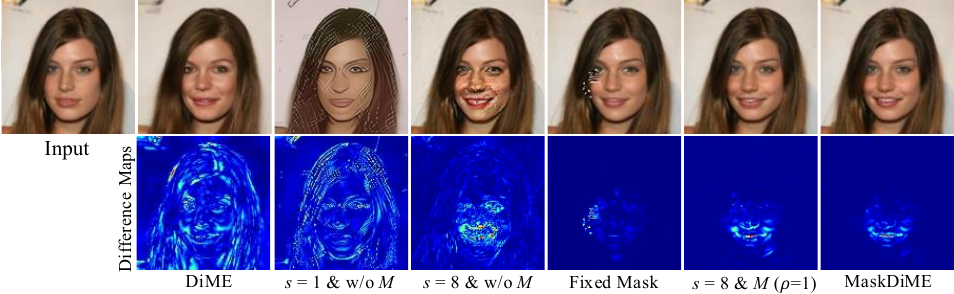}

\caption{\textbf{Ablation study on the \textit{No Smile} $\rightarrow$ \textit{Smile} sample of CelebA.} 
  Increasing the gradient scaling ($s$) and introducing Adaptive Dual-mask ($M$) progressively improve MaskDiME’s results, localizing edits to decision-relevant regions and yielding realistic, semantically consistent counterfactuals.
}

   \label{fig:ablation}
\end{figure}

\begin{table}[t]
  \centering
  \caption{Ablation study of MaskDiME on \textit{Smile} of CelebA.}
  \label{tab:maskdime_ablation}
  \fontsize{9pt}{10pt}\selectfont
  \setlength{\tabcolsep}{1.6pt}
  \begin{tabular}{lcccccc}
    \toprule
    Method & FID↓ & FS↑ & MNAC↓ & CD↓ & COUT↑ & FR(\%)↑ \\
    \midrule
    DiME      & 3.17& 0.73 & 3.72 & 2.30 & 0.53 & 97.2 \\
    \midrule

   $s$=1 \& w/o mask     & 95.76 & 0.63  & 6.15 & 2.43 &  -0.16& 55.2\\
    % $s$=1 and  mask     &  &  & &  &  & \\
    $s$=8 \& w/o mask     & 15.94  & 0.77  & 5.71& 4.20 & 0.96 & 100.0\\
 Fixed Mask    & 4.21 & 0.86  & 2.98 & 2.03 & 0.70&99.7\\
    $s$=8 \& mask ($\rho$=1)     & 0.71 & 0.90  & 2.66  &  2.25&0.81  &100.0 \\
    MaskDiME ($\rho$=0.5)    & 0.71  & 0.91 & 2.78 & 2.41 & 0.87 &100.0 \\
    \bottomrule
  \end{tabular}
\end{table}

\noindent \noindent \textbf{Ablation Study.} 
We conduct an ablation study to analyze the key components of MaskDiME, as shown in \cref{fig:ablation,tab:maskdime_ablation}. 
The main difference between DiME and the ``$s=1$ \& w/o mask'' variant lies in how the clean image is obtained: DiME reconstructs it recursively through the denoising diffusion process, whereas the latter directly estimates it in a single step using \cref{eq:get_clean}. 
This fast one-step estimation leads to a substantial performance degradation, with much higher FID (95.76), lower COUT (-0.16), and reduced FR (55.2\%). Increasing the gradient scaling factor to $s=8$ amplifies the classifier, perceptual, and $L_1$ loss gradients, producing visually more realistic images and successfully flipping the classifier prediction (FR = 100\%). 
However, the resulting strong and unconstrained gradients degrade image quality and semantic consistency, yielding higher FID (15.94), and visible artifacts. 
By contrast, a fixed mask suppresses irrelevant edits to some extent but lacks adaptability to sample- and gradient-level variations, limiting the trade-off between image quality and decision consistency. Introducing the adaptive dual-mask ($\rho=1$) localizes edits to decision-relevant regions, producing more natural counterfactuals. 
% As a result, FID improves from 15.94 to 0.71 and FS from 0.77 to 0.90, while FR remains at 100\%. 
Finally, the full MaskDiME configuration further refines the clean-image blending ratio ($\rho=0.5$), maintaining a similar FID (0.71) while improving decision consistency (COUT from 0.81 to 0.87). 
Overall, gradient scaling and masking complement each other: the former controls guidance strength, while the latter enforces spatially focused and semantically coherent edits.

\noindent \textbf{Limitations} Although MaskDiME significantly improves the efficiency and performance of diffusion-based counterfactual explanations, it still exhibits certain inherent limitations.
First, MaskDiME relies on classifier gradients, which may cause inaccurate mask localization on multi-class datasets (e.g., ImageNet) due to noisy and spatially inconsistent gradient signals. An additional mask extractor may mitigate this problem but at the cost of higher computation and lower flexibility. Second, current methods, including MaskDiME, generally lack ground-truth annotations for counterfactual explanations, making it difficult to validate their results from a causal perspective. Finally, the evaluation of MaskDiME still depends on perceptual similarity metrics such as FID, which may introduce bias when large unmodified regions exist in the image.

\section{Conclusions}
We propose MaskDiME, a training-free diffusion framework that introduces an adaptive dual-mask mechanism guided by classifier gradients for spatially precise counterfactual generation. Our experiments demonstrate that simple adjustments to the sampling process, such as employing dynamic masks to constrain the update regions, enable MaskDiME to maintain high generation efficiency and achieve state-of-the-art performance across multiple visual domains while minimizing GPU memory usage. These findings highlight the importance of spatially adaptive constraints in diffusion-based explainability, providing a practical approach to more efficient and precise visual counterfactual explanations.

{
    \small

    \bibliographystyle{ieeenat_fullname}
    
    \bibliography{main}

@String(CVPR= {IEEE Conf. Comput. Vis. Pattern Recog.})

@String(ICLR = {Int. Conf. Learn. Represent.})

@String(CVPR  = {CVPR})

@String(ICLR  = {ICLR})

@inproceedings{goodfellow2015explaining,
  title     = {Explaining and Harnessing Adversarial Examples},
  author    = {Goodfellow, Ian J. and Shlens, Jonathon and Szegedy, Christian},
  booktitle = {Proceedings of the 3rd International Conference on Learning Representations (ICLR)},
  year      = {2015},
  address   = {San Diego, CA, USA},
  editor    = {Bengio, Yoshua and LeCun, Yann}
}

@inproceedings{liu2015deep,
  title={Deep learning face attributes in the wild},
  author={Liu, Ziwei and Luo, Ping and Wang, Xiaogang and Tang, Xiaoou},
  booktitle={Proceedings of the IEEE international conference on computer vision},
  pages={3730--3738},
  year={2015}
}

@inproceedings{lee2020maskgan,
  title={Maskgan: Towards diverse and interactive facial image manipulation},
  author={Lee, Cheng-Han and Liu, Ziwei and Wu, Lingyun and Luo, Ping},
  booktitle={Proceedings of the IEEE/CVF conference on computer vision and pattern recognition},
  pages={5549--5558},
  year={2020}
}

@inproceedings{jeanneret2022diffusion,
  title={Diffusion models for counterfactual explanations},
  author={Jeanneret, Guillaume and Simon, Lo{\"\i}c and Jurie, Fr{\'e}d{\'e}ric},
  booktitle={Proceedings of the Asian conference on computer vision},
  pages={858--876},
  year={2022}
}

@inproceedings{jeanneret2023adversarial,
  title={Adversarial counterfactual visual explanations},
  author={Jeanneret, Guillaume and Simon, Lo{\"\i}c and Jurie, Fr{\'e}d{\'e}ric},
  booktitle={Proceedings of the IEEE/CVF Conference on Computer Vision and Pattern Recognition},
  pages={16425--16435},
  year={2023}
}

@inproceedings{yu2020bdd100k,
  title={Bdd100k: A diverse driving dataset for heterogeneous multitask learning},
  author={Yu, Fisher and Chen, Haofeng and Wang, Xin and Xian, Wenqi and Chen, Yingying and Liu, Fangchen and Madhavan, Vashisht and Darrell, Trevor},
  booktitle={Proceedings of the IEEE/CVF conference on computer vision and pattern recognition},
  pages={2636--2645},
  year={2020}
}

@inproceedings{deng2009imagenet,
  title={Imagenet: A large-scale hierarchical image database},
  author={Deng, Jia and Dong, Wei and Socher, Richard and Li, Li-Jia and Li, Kai and Fei-Fei, Li},
  booktitle={2009 IEEE conference on computer vision and pattern recognition},
  pages={248--255},
  year={2009},
  organization={Ieee}
}

@inproceedings{jacob2022steex,
  title={STEEX: steering counterfactual explanations with semantics},
  author={Jacob, Paul and Zablocki, {\'E}loi and Ben-Younes, Hedi and Chen, Micka{\"e}l and P{\'e}rez, Patrick and Cord, Matthieu},
  booktitle={European Conference on Computer Vision},
  pages={387--403},
  year={2022},
  organization={Springer}
}

@inproceedings{xu2020explainable,
  title={Explainable object-induced action decision for autonomous vehicles},
  author={Xu, Yiran and Yang, Xiaoyin and Gong, Lihang and Lin, Hsuan-Chu and Wu, Tz-Ying and Li, Yunsheng and Vasconcelos, Nuno},
  booktitle={Proceedings of the IEEE/CVF Conference on Computer Vision and Pattern Recognition},
  pages={9523--9532},
  year={2020}
}

@inproceedings{weng2024fast,
  title={Fast diffusion-based counterfactuals for shortcut removal and generation},
  author={Weng, Nina and Pegios, Paraskevas and Petersen, Eike and Feragen, Aasa and Bigdeli, Siavash},
  booktitle={European Conference on Computer Vision},
  pages={338--357},
  year={2024},
  organization={Springer}
}

@article{farid2023latent,
  title={Latent diffusion counterfactual explanations},
  author={Farid, Karim and Schrodi, Simon and Argus, Max and Brox, Thomas},
  journal={arXiv preprint arXiv:2310.06668},
  year={2023}
}

@inproceedings{jeanneret2024text,
  title={Text-to-image models for counterfactual explanations: a black-box approach},
  author={Jeanneret, Guillaume and Simon, Lo{\"\i}c and Jurie, Fr{\'e}d{\'e}ric},
  booktitle={Proceedings of the IEEE/CVF Winter Conference on Applications of Computer Vision},
  pages={4757--4767},
  year={2024}
}

@inproceedings{sobieski2024rethinking,
  title={Rethinking visual counterfactual explanations through region constraint},
  author={Sobieski, Bartlomiej and Grzywaczewski, Jakub and Sadlej, Bart{\l}omiej and Tivnan, Matthew and Biecek, Przemyslaw},
  booktitle={The Thirteenth International Conference on Learning Representations},
  year={2025}
}

@article{heusel2017gans,
  title={Gans trained by a two time-scale update rule converge to a local nash equilibrium},
  author={Heusel, Martin and Ramsauer, Hubert and Unterthiner, Thomas and Nessler, Bernhard and Hochreiter, Sepp},
  journal={Advances in neural information processing systems},
  volume={30},
  year={2017}
}

@inproceedings{cao2018vggface2,
  title={Vggface2: A dataset for recognising faces across pose and age},
  author={Cao, Qiong and Shen, Li and Xie, Weidi and Parkhi, Omkar M and Zisserman, Andrew},
  booktitle={2018 13th IEEE international conference on automatic face \& gesture recognition (FG 2018)},
  pages={67--74},
  year={2018},
  organization={IEEE}
}

@inproceedings{chen2021exploring,
  title={Exploring simple siamese representation learning},
  author={Chen, Xinlei and He, Kaiming},
  booktitle={Proceedings of the IEEE/CVF conference on computer vision and pattern recognition},
  pages={15750--15758},
  year={2021}
}

@inproceedings{khorram2022cycle,
  title={Cycle-consistent counterfactuals by latent transformations},
  author={Khorram, Saeed and Fuxin, Li},
  booktitle={Proceedings of the IEEE/CVF Conference on Computer Vision and Pattern Recognition},
  pages={10203--10212},
  year={2022}
}

@inproceedings{yeganeh2025latent,
  title={Latent Drifting in Diffusion Models for Counterfactual Medical Image Synthesis},
  author={Yeganeh, Yousef and Farshad, Azade and Charisiadis, Ioannis and Hasny, Marta and Hartenberger, Martin and Ommer, Bj{"o}rn and Navab, Nassir and Adeli, Ehsan},
  booktitle={Proceedings of the IEEE/CVF Conference on Computer Vision and Pattern Recognition (CVPR)},
  year={2025}
}

@article{ho2020denoising,
  title={Denoising diffusion probabilistic models},
  author={Ho, Jonathan and Jain, Ajay and Abbeel, Pieter},
  journal={Advances in neural information processing systems},
  volume={33},
  pages={6840--6851},
  year={2020}
}

@inproceedings{kazimi2025explaining,
  title={Explaining in Diffusion: Explaining a Classifier with Diffusion Semantics},
  author={Kazimi, Tahira and Allada, Ritika and Yanardag, Pinar},
  booktitle={Proceedings of the Computer Vision and Pattern Recognition Conference},
  pages={14799--14809},
  year={2025}
}

@inproceedings{zhu2025interpretable,
  title={Interpretable Image Classification via Non-parametric Part Prototype Learning},
  author={Zhu, Zhijie and Fan, Lei and Pagnucco, Maurice and Song, Yang},
  booktitle={Proceedings of the Computer Vision and Pattern Recognition Conference},
  pages={9762--9771},
  year={2025}
}

@inproceedings{singh2025protopatchnet,
  title={ProtoPatchNet: An Interpretable Patch-Based Prototypical Network},
  author={Singh, Mohana and Gubbi, Jayavardhana and Babu, R Venkatesh and others},
  booktitle={Proceedings of the Computer Vision and Pattern Recognition Conference},
  pages={721--728},
  year={2025}
}

@inproceedings{kapse2024si,
  title={Si-mil: Taming deep mil for self-interpretability in gigapixel histopathology},
  author={Kapse, Saarthak and Pati, Pushpak and Das, Srijan and Zhang, Jingwei and Chen, Chao and Vakalopoulou, Maria and Saltz, Joel and Samaras, Dimitris and Gupta, Rajarsi R and Prasanna, Prateek},
  booktitle={Proceedings of the IEEE/CVF Conference on Computer Vision and Pattern Recognition},
  pages={11226--11237},
  year={2024}
}

@inproceedings{disciple2025learning,
  title={Disciple: Learning interpretable programs for scientific visual discovery},
  author={Mall, Utkarsh and Phoo, Cheng Perng and Chiquier, Mia and Hariharan, Bharath and Bala, Kavita and Vondrick, Carl},
  booktitle={Proceedings of the Computer Vision and Pattern Recognition Conference},
  pages={29258--29267},
  year={2025}
}

@inproceedings{zhang2025towards,
  title={Towards Fine-Grained Interpretability: Counterfactual Explanations for Misclassification with Saliency Partition},
  author={Zhang, Lintong and Yin, Kang and Lee, Seong-Whan},
  booktitle={Proceedings of the Computer Vision and Pattern Recognition Conference},
  pages={30053--30062},
  year={2025}
}

@inproceedings{huy2025interactive,
  title={Interactive Medical Image Analysis with Concept-based Similarity Reasoning},
  author={Huy, Ta Duc and Tran, Sen Kim and Nguyen, Phan and Tran, Nguyen Hoang and Sam, Tran Bao and van den Hengel, Anton and Liao, Zhibin and Verjans, Johan W and To, Minh-Son and Phan, Vu Minh Hieu},
  booktitle={Proceedings of the Computer Vision and Pattern Recognition Conference},
  pages={30797--30806},
  year={2025}
}

@inproceedings{zhang2018interpretable,
  title={Interpretable convolutional neural networks},
  author={Zhang, Quanshi and Wu, Ying Nian and Zhu, Song-Chun},
  booktitle={Proceedings of the IEEE conference on computer vision and pattern recognition},
  pages={8827--8836},
  year={2018}
}

@article{rudin2019stop,
  title={Stop explaining black box machine learning models for high stakes decisions and use interpretable models instead},
  author={Rudin, Cynthia},
  journal={Nature machine intelligence},
  volume={1},
  number={5},
  pages={206--215},
  year={2019},
  publisher={Nature Publishing Group UK London}
}

@inproceedings{zheng2022shap,
  title={Shap-CAM: Visual explanations for convolutional neural networks based on Shapley value},
  author={Zheng, Quan and Wang, Ziwei and Zhou, Jie and Lu, Jiwen},
  booktitle={European conference on computer vision},
  pages={459--474},
  year={2022},
  organization={Springer}
}

@inproceedings{selvaraju2017grad,
  title={Grad-cam: Visual explanations from deep networks via gradient-based localization},
  author={Selvaraju, Ramprasaath R and Cogswell, Michael and Das, Abhishek and Vedantam, Ramakrishna and Parikh, Devi and Batra, Dhruv},
  booktitle={Proceedings of the IEEE international conference on computer vision},
  pages={618--626},
  year={2017}
}

@inproceedings{jalwana2021cameras,
  title={Cameras: Enhanced resolution and sanity preserving class activation mapping for image saliency},
  author={Jalwana, Mohammad AAK and Akhtar, Naveed and Bennamoun, Mohammed and Mian, Ajmal},
  booktitle={Proceedings of the IEEE/CVF Conference on Computer Vision and Pattern Recognition},
  pages={16327--16336},
  year={2021}
}

@inproceedings{chen2025explainable,
  title={Explainable Saliency: Articulating Reasoning with Contextual Prioritization},
  author={Chen, Nuo and Jiang, Ming and Zhao, Qi},
  booktitle={Proceedings of the Computer Vision and Pattern Recognition Conference},
  pages={9601--9610},
  year={2025}
}

@inproceedings{kim2018interpretability,
  title={Interpretability beyond feature attribution: Quantitative testing with concept activation vectors (tcav)},
  author={Kim, Been and Wattenberg, Martin and Gilmer, Justin and Cai, Carrie and Wexler, James and Viegas, Fernanda and others},
  booktitle={International conference on machine learning},
  pages={2668--2677},
  year={2018},
  organization={PMLR}
}

@article{ghorbani2019towards,
  title={Towards automatic concept-based explanations},
  author={Ghorbani, Amirata and Wexler, James and Zou, James Y and Kim, Been},
  journal={Advances in neural information processing systems},
  volume={32},
  year={2019}
}

@inproceedings{yu2025coe,
  title={CoE: Chain-of-Explanation via Automatic Visual Concept Circuit Description and Polysemanticity Quantification},
  author={Yu, Wenlong and Wang, Qilong and Liu, Chuang and Li, Dong and Hu, Qinghua},
  booktitle={Proceedings of the Computer Vision and Pattern Recognition Conference},
  pages={4364--4374},
  year={2025}
}

@inproceedings{grobrugge2025towards,
  title={Towards Human-Understandable Multi-Dimensional Concept Discovery},
  author={Grobr{\"u}gge, Arne and K{\"u}hl, Niklas and Satzger, Gerhard and Spitzer, Philipp},
  booktitle={Proceedings of the Computer Vision and Pattern Recognition Conference},
  pages={20018--20027},
  year={2025}
}

@article{tan2018learning,
  title={Learning global additive explanations for neural nets using model distillation},
  author={Tan, Sarah and Caruana, Rich and Hooker, Giles and Koch, Paul and Gordo, Albert},
  journal={stat},
  volume={1050},
  number={3},
  pages={1518--68323},
  year={2018}
}

@inproceedings{haselhoff2021towards,
  title={Towards black-box explainability with Gaussian discriminant knowledge distillation},
  author={Haselhoff, Anselm and Kronenberger, Jan and Kuppers, Fabian and Schneider, Jonas},
  booktitle={Proceedings of the IEEE/CVF Conference on Computer Vision and Pattern Recognition},
  pages={21--28},
  year={2021}
}

@inproceedings{haselhoff2024gaussian,
  title={The gaussian discriminant variational autoencoder (gdvae): A self-explainable model with counterfactual explanations},
  author={Haselhoff, Anselm and Trelenberg, Kevin and K{\"u}ppers, Fabian and Schneider, Jonas},
  booktitle={European Conference on Computer Vision},
  pages={305--322},
  year={2024},
  organization={Springer}
}

@inproceedings{lang2021explaining,
  title={Explaining in style: training a GAN to explain a classifier in stylespace},
  author={Lang, Oran and Gandelsman, Yossi and Yarom, Michal and Wald, Yoav and Elidan, Gal and Hassidim, Avinatan and Freeman, William T and Isola, Phillip and Globerson, Amir and Irani, Michal and others},
  booktitle={Proceedings of the IEEE/CVF International Conference on Computer Vision},
  pages={693--702},
  year={2021}
}

@article{efron2011tweedie,
  title={Tweedie’s formula and selection bias},
  author={Efron, Bradley},
  journal={Journal of the American Statistical Association},
  volume={106},
  number={496},
  pages={1602--1614},
  year={2011},
  publisher={Taylor \& Francis}
}

@article{song2020score,
  title={Score-based generative modeling through stochastic differential equations},
  author={Song, Yang and Sohl-Dickstein, Jascha and Kingma, Diederik P and Kumar, Abhishek and Ermon, Stefano and Poole, Ben},
  journal={arXiv preprint arXiv:2011.13456},
  year={2020}
}

@inproceedings{rombach2022high,
  title={High-resolution image synthesis with latent diffusion models},
  author={Rombach, Robin and Blattmann, Andreas and Lorenz, Dominik and Esser, Patrick and Ommer, Bj{\"o}rn},
  booktitle={Proceedings of the IEEE/CVF conference on computer vision and pattern recognition},
  pages={10684--10695},
  year={2022}
}

@article{liu20232,
  title = {I$^{2}$SB: Image-to-Image Schr{\"o}dinger Bridge},
  author = {Liu, Guan-Horng and Vahdat, Arash and Huang, De-An and Theodorou, Evangelos A. and Nie, Weili and Anandkumar, Anima},
  journal = {arXiv preprint arXiv:2302.05872},
  year = {2023}
}

@inproceedings{ronneberger2015u,
  title={U-net: Convolutional networks for biomedical image segmentation},
  author={Ronneberger, Olaf and Fischer, Philipp and Brox, Thomas},
  booktitle={International Conference on Medical image computing and computer-assisted intervention},
  pages={234--241},
  year={2015},
  organization={Springer}
}

@inproceedings{rodriguez2021beyond,
  title={Beyond trivial counterfactual explanations with diverse valuable explanations},
  author={Rodriguez, Pau and Caccia, Massimo and Lacoste, Alexandre and Zamparo, Lee and Laradji, Issam and Charlin, Laurent and Vazquez, David},
  booktitle={Proceedings of the IEEE/CVF International Conference on Computer Vision},
  pages={1056--1065},
  year={2021}
}

@article{dhariwal2021diffusion,
  title={Diffusion models beat gans on image synthesis},
  author={Dhariwal, Prafulla and Nichol, Alexander},
  journal={Advances in neural information processing systems},
  volume={34},
  pages={8780--8794},
  year={2021}
}

@inproceedings{sundararajan2017axiomatic,
  title={Axiomatic attribution for deep networks},
  author={Sundararajan, Mukund and Taly, Ankur and Yan, Qiqi},
  booktitle={International conference on machine learning},
  pages={3319--3328},
  year={2017},
  organization={PMLR}
}

@article{wachter2017counterfactual,
  title={Counterfactual explanations without opening the black box: Automated decisions and the GDPR},
  author={Wachter, Sandra and Mittelstadt, Brent and Russell, Chris},
  journal={Harv. JL \& Tech.},
  volume={31},
  pages={841},
  year={2017},
  publisher={HeinOnline}
}

@article{augustin2022diffusion,
  title={Diffusion visual counterfactual explanations},
  author={Augustin, Maximilian and Boreiko, Valentyn and Croce, Francesco and Hein, Matthias},
  journal={Advances in Neural Information Processing Systems},
  volume={35},
  pages={364--377},
  year={2022}
}

@article{smilkov2017smoothgrad,
  title={Smoothgrad: removing noise by adding noise},
  author={Smilkov, Daniel and Thorat, Nikhil and Kim, Been and Vi{\'e}gas, Fernanda and Wattenberg, Martin},
  journal={arXiv preprint arXiv:1706.03825},
  year={2017}
}

@article{simonyan2013deep,
  title={Deep inside convolutional networks: Visualising image classification models and saliency maps},
  author={Simonyan, Karen and Vedaldi, Andrea and Zisserman, Andrew},
  journal={arXiv preprint arXiv:1312.6034},
  year={2013}
}

@inproceedings{chowdhury2025looking,
  title={Looking in the Mirror: A Faithful Counterfactual Explanation Method for Interpreting Deep Image Classification Models},
  author={Chowdhury, Townim and Phan, Vu Minh Hieu and Liao, Kewen and Dong, Nanyu and To, Minh-Son and van den Hengel, Anton and Verjans, Johan W and Liao, Zhibin},
  booktitle={Proceedings of the IEEE/CVF International Conference on Computer Vision},
  pages={2239--2249},
  year={2025}
}

@inproceedings{zhang2025finer,
  title={Finer-cam: Spotting the difference reveals finer details for visual explanation},
  author={Zhang, Ziheng and Gu, Jianyang and Chowdhury, Arpita and Mai, Zheda and Carlyn, David and Berger-Wolf, Tanya and Su, Yu and Chao, Wei-Lun},
  booktitle={Proceedings of the Computer Vision and Pattern Recognition Conference},
  pages={9611--9620},
  year={2025}
}

@inproceedings{zhang2018unreasonable,
  title={The unreasonable effectiveness of deep features as a perceptual metric},
  author={Zhang, Richard and Isola, Phillip and Efros, Alexei A and Shechtman, Eli and Wang, Oliver},
  booktitle={Proceedings of the IEEE conference on computer vision and pattern recognition},
  pages={586--595},
  year={2018}
}
}

% \clearpage
% \clearpage
% {
%     \small

%     \bibliographystyle{ieeenat_fullname}
    
%     \bibliography{main}
% }

\end{document}